\documentclass[10pt,journal,compsoc]{IEEEtran}

\selectfont

\usepackage[english]{babel} 
\usepackage{amssymb}
\usepackage{amsmath}
\usepackage{mathdots}
\usepackage{graphicx}
\usepackage{authblk}
\usepackage{diagbox}
\PassOptionsToPackage{hyphens}{url}\usepackage{hyperref}

\ifCLASSOPTIONcompsoc
  \usepackage[nocompress]{cite}
\else
  \usepackage{cite}
\fi
\ifCLASSINFOpdf
\else
\fi
\begin{document}
%
\title{WonDerM: Skin Lesion Classification with Fine-tuned Neural Networks}

%
%
%
%

\author[1]{Yeong Chan~Lee}
\author[1]{Sang-Hyuk~Jung}

\author[1$\dagger$]{Hong-Hee~Won\thanks{ * Yeong Chan Lee and Sang-Hyuk Jung contributed equally to this work.}}




%
%

\markboth{ISIC 2018:Skin Lesion Analysis Towards Melanoma Detection}%
{Yeong Chan~Lee, Sang-Hyuk~Jung, Hong-Hee~Won \MakeLowercase{\textit{et al.}}: WonDerM: Skin Lesion Classification with Fine-tuned Neural Networks}
%



\affil[1]{
	Samsung Advanced Institute of Health Sciences and Technology (SAIHST), Sungkyunkwan University, Samsung Medical Center, Seoul, Republic of Korea
}
\affil[$\dagger$]{
	Correspondence to ${honghee.won}$ at ${gmail.com}$
}

\IEEEtitleabstractindextext{%
\begin{abstract}
As skin cancer is one of the most frequent cancers globally, accurate, non-invasive dermoscopy-based diagnosis becomes essential and promising. A task of the Part 3 of the ISIC Skin Image Analysis Challenge@MICCAI 2018 is to predict seven disease classes with skin lesion images, including melanoma (MEL), melanocytic nevus (NV), basal cell carcinoma (BCC), actinic keratosis / Bowen’s disease (intraepithelial carcinoma) (AKIEC), benign keratosis (solar lentigo / seborrheic keratosis / lichen planus-like keratosis) (BKL), dermatofibroma (DF) and vascular lesion (VASC) as defined by the International Dermatology Society \cite{test_1},\cite{test}.
In this work, we design the WonDerM pipeline, that resamples the preprocessed skin lesion images, builds neural network architecture fine-tuned with segmentation task data (the Part 1), and uses an ensemble method to classify the seven skin diseases. Our model achieved an accuracy of 0.899 and 0.785 in the validation set and test set, respectively.

\end{abstract}

}

\maketitle

\IEEEdisplaynontitleabstractindextext

%
\IEEEpeerreviewmaketitle

\ifCLASSOPTIONcompsoc
\IEEEraisesectionheading{\section{Introduction}\label{sec:introduction}}
\else
\section{Introduction}
\label{sec:introduction}
\fi

%
%
%
%
\IEEEPARstart {S}{kin} cancer is one of the most common cancers that has been increasing worldwide \cite{3}. There are several types of skin cancer, including melanoma, basal cell carcinoma, squamous cell carcinoma, intraepithelial carcinoma, etc. \cite{4},\cite{5},\cite{6}. Among them, melanoma is the highest mortality rate cancer and has been primarily studied in many previous studies compared with other skin cancer \cite{7}.

Dermatologists established the classification system for an accurate diagnosis of skin disorders. Because there are diverse characteristics of skin cancer, similarity to benign lesions and specific lesions seen from diseases, distinguishing skin cancer from other skin disorders is important in dermatology clinics. Although skin cancer can be early detected by direct interference, such visual similarity of lesions and various patterns make it difficult to diagnose the exact type of cancer \cite{8}. Therefore, in the past few years, dermoscopy has been enabled to generate unprecedentedly large amount of detailed images of skin cancer and other skin diseases and dermoscopy-based non-invasive diagnosis has been developed rapidly \cite{9},\cite{10},\cite{11}.

Recently, deep learning approaches have improved prediction performance in classification and diagnosis of diseases based on medical imaging. The approaches have been applied not only for predicting the presence of disease but also for distinguishing the classes of disease \cite{12}. In particular, recent studies demonstrated remarkable prediction performance in classifying skin cancer using deep learning algorithm in binary classification \cite{13}. Multi-class classification model did not achieve comparable performance to the binary classification model.

In this work, we propose the WonDerM pipeline that is optimized to predict multiple skin disease classes as shown in Figure 1. First, the preprocessed images are divided into the training set and development set. The images in the training set are resampled for balancing. Second, the combined model (DenseNet and U-net) is trained for segmentation based on the ISIC Skin Image Analysis Challenge segmentation (Part 1) data and used for fine-tuning of the subsequent classification models. Third, the extracted architecture (encoder part) from the segmentation model is further trained for classifying the seven skin diseases based on the preprocessed ISIC Skin Image Analysis Challenge classification (Part 3) data. Last, an ensemble method of multiple trained classifiers is applied to predict the disease class.


\begin{figure*}
	\centering
	\includegraphics[width=17.5cm]{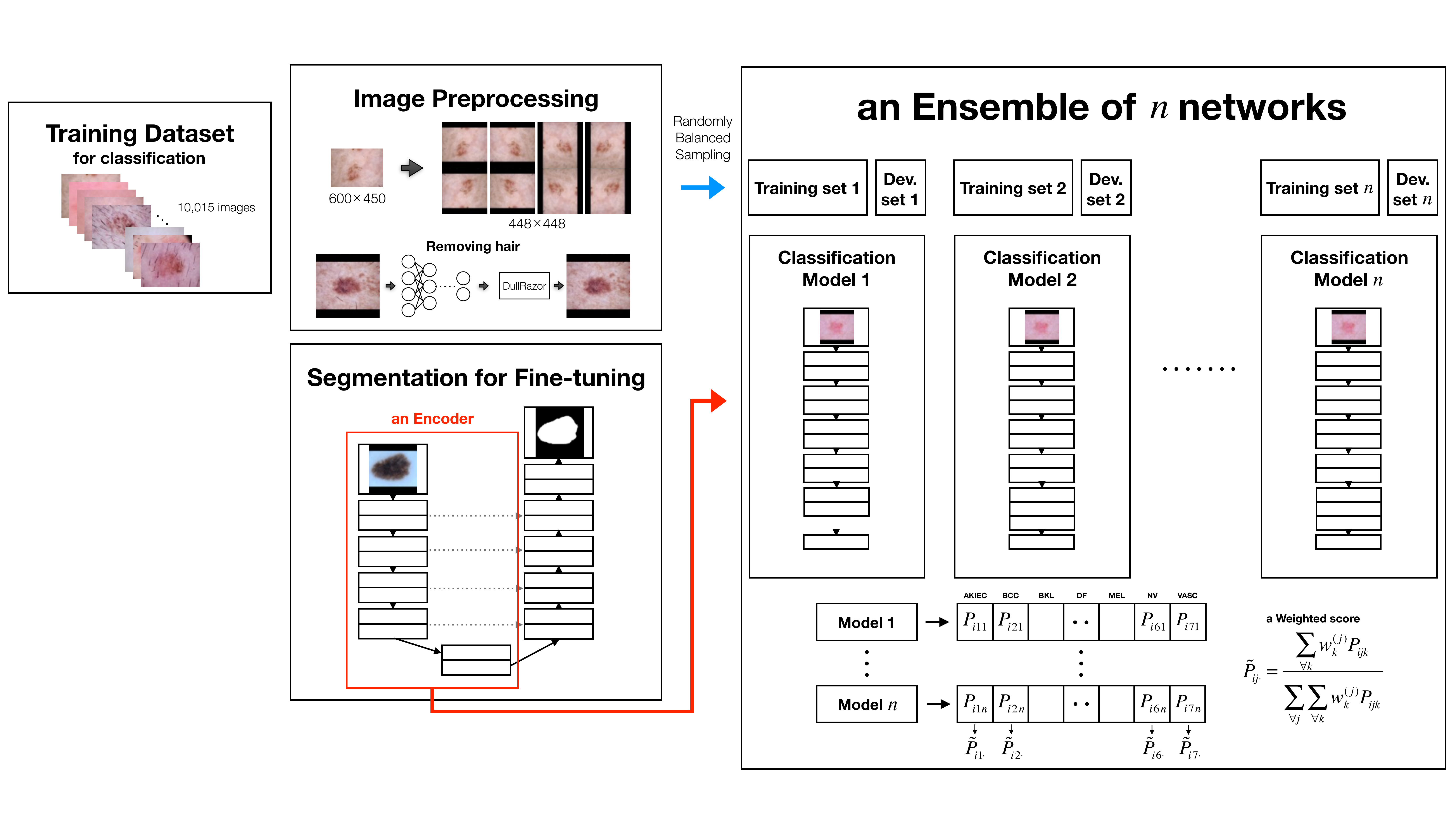}
	\caption{The WonDerM pipeline}
\end{figure*}

\section{Method}

\subsection{Datasets} Dermoscopic lesion images were acquired from HAM10000 Dataset via several institutions \cite{test},\cite{14}. A disease label for each image was determined histopathologically or diagnostically. A training dataset for classification consisted of 10,015 images (327 AKIEC, 514 BCC, 1,099 BKL, 115 DF, 1,113 MEL, 6,705 NV, and 142 VASC samples) with the corresponding disease label (ground truth), and a validation set and a test set comprised 193 images and 1,512 images without ground truth, respectively. The segmentation dataset for fine-tuning is made of 2,594 images with corresponding masks.

\subsection{Image Preprocessing} The original images with 450${\times}$600 pixels were padded to 600${\times}$600 pixels for resolution and resized to 448${\times}$448 for DenseNet. In the skin lesion analysis, hairs on the skin might interfere the lesion of interest, which can be a noise. We developed a simple convolutional neural network to classify the skin image with hairs or without hairs. Hairs of the classified skin images were removed \cite{15}.

For data augmentation, the images were rotated by 90, 180, and 270 degrees, and flipped vertically. In addition, we flipped images of two classes (DF and VASC) horizontally that were relatively fewer than others to avoid imbalance among classes of each resampled dataset (Table 1). Finally, we obtained 8 or 12 images per image, resulting in a total of 28,052 images.

\subsection{Fine-tuning with Higher-order Task Data} Inductive transfer learning is an approach that the target task takes knowledge of source domain and source task when the source task is different from the target task \cite{12}. Then, the learning of target task can be improved if the knowledge in the source domain is more than in the task domain.

In general, segmentation task has much information than classification task. The data for segmentation basically have pixel-wise information which is spatial and morphological. When learning the data, the model may learn skin lesion features which it needs to focus on. Thus, our model was pre-trained with the segmentation dataset so that it could learn morphological features although the data are not labelled with skin diseases.

Our neural network for segmentation was composed of DenseNet combined with U-net architecture \cite{16}. Figure 2. describes the segmentation architecture in which DenseNet is used for an encoder and U-net for a decoder. The convolution layer right before the first dense block and the last concatenated convolutions of each dense block of DenseNet are skip-connected with the convolutions of U-net.

\begin{figure}
	\centering
	\includegraphics[width=8cm]{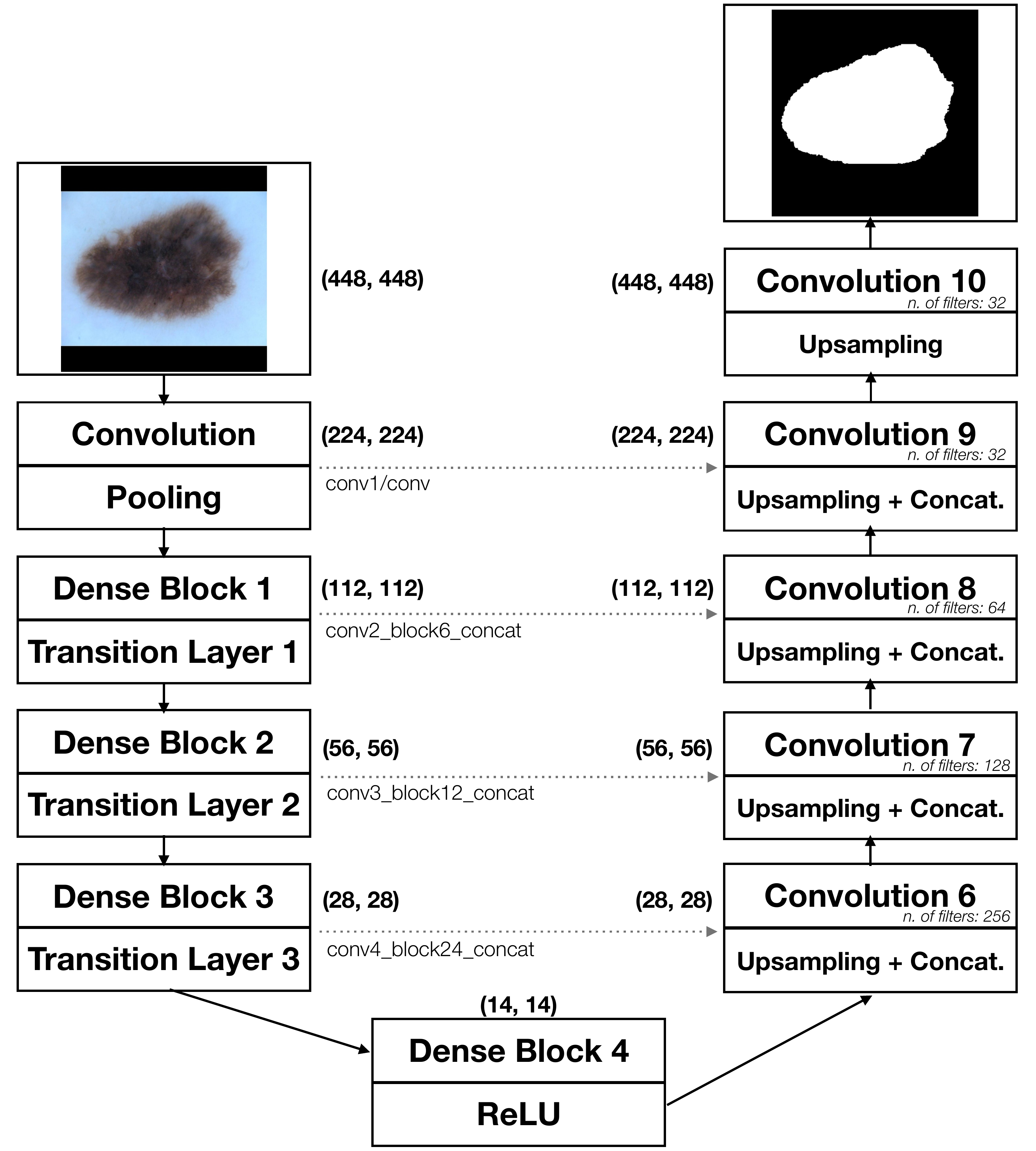}
	\caption{The network architecture for segmentation}
\end{figure}

\subsection{Classification Model Architecture} We used DenseNet model architecture pre-trained with segmentation data. The model was extracted from the encoder part of the segmentation model. As shown in Figure 3, it was added to convolution layer, global average pooling layer, and classification layer after the last dense block. Our model has approximately 16 million parameters.

\begin{figure}
	\centering
	\includegraphics[width=5cm]{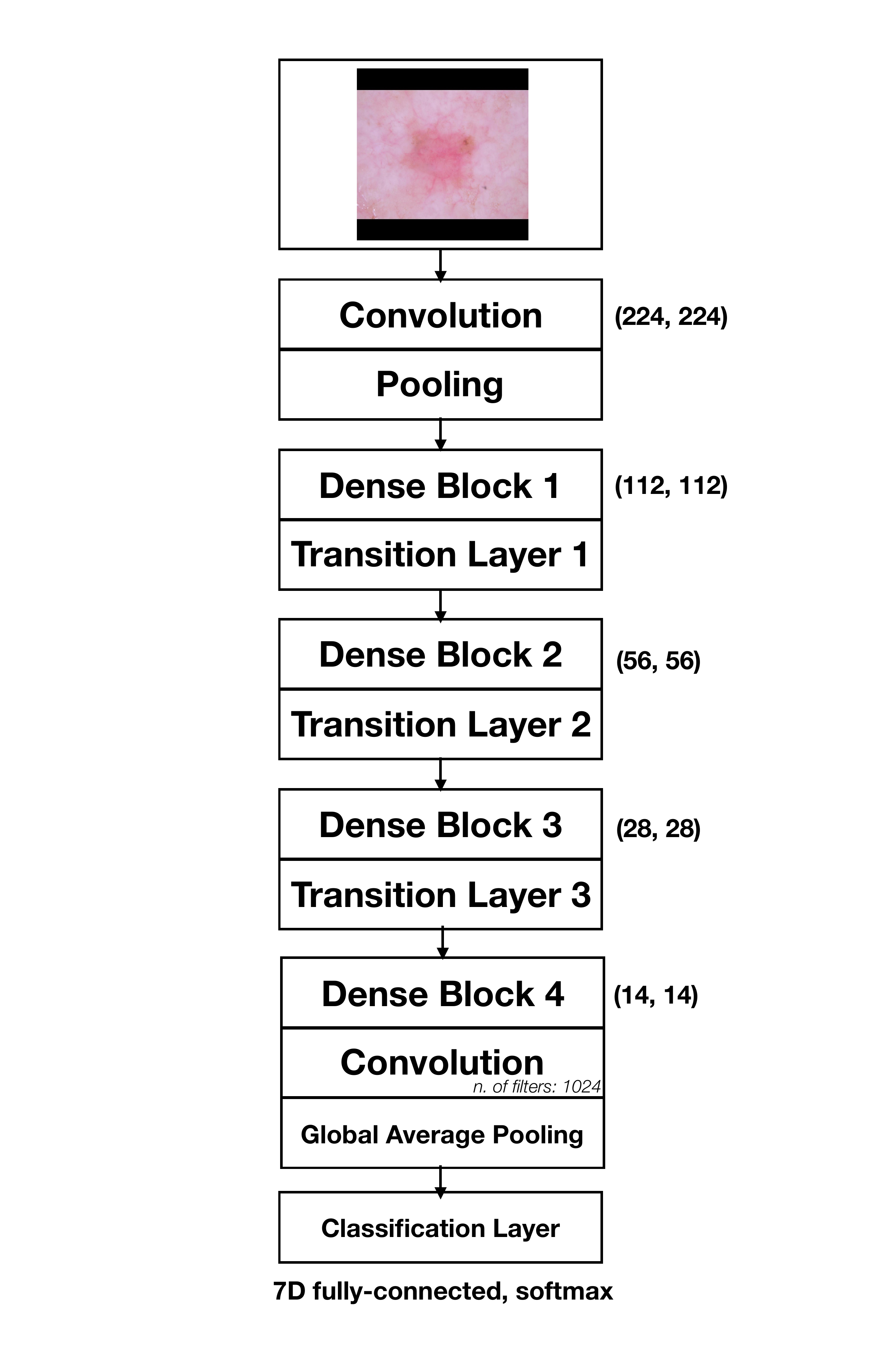}
	\caption{The network architecture for classification}
\end{figure}

\subsection{Randomly Balanced Sampling} The training images of skin disease classes were highly imbalanced (Table 1). For instance, the number of images of the DF class was approximately 60 times smaller than the NV class. To minimize the imbalance in the training set, first, the given data was again divided into the training set for learning skin lesions and the development set for tuning the model at a 9:1 ratio. Then, only the training data were resampled for balancing the number of images per class. The number of images in all the classes was fitted to the number of images in certain selected classes with fewer images.

We generated 4 balanced datasets for an ensemble in this way. We fitted the number of images in classes to the number of BCC images (nearly 463 images).

\begin{table}[]
	\begin{tabular}{lccc}
		\hline
		 & \begin{tabular}[c]{@{}c@{}} Original training set \end{tabular} & \begin{tabular}[c]{@{}c@{}} Training set\\{(1-4)} \end{tabular} & \begin{tabular}[c]{@{}c@{}} Development set\\{(1-4)} \end{tabular} \\ [1.5ex] \hline
		
		{AKIEC} & 327     & 294   & 33   \\ [1.08ex]
		{BCC}   & 514     & 463   & 51   \\ [1.08ex]   
		{BKL}   & 1,099   & 463   & 110  \\ [1.08ex]   
		{DF}    & 115     & 103   & 12   \\ [1.08ex]   
		{MEL}   & 1,113   & 463   & 111  \\ [1.08ex]   
		{NV}    & 6,705   & 463   & 670  \\ [1.08ex]   
		{VASC}  & 142     & 128   & 14   \\  \hline 
		
		\end{tabular}
		\begin{tabular}[c]{@{}c@{}} \\ Table 1. The number of images in the original training dataset, the \\ resampled balanced training dataset, and the development dataset.\end{tabular}

\end{table}


\subsection{Ensemble of Networks} An ensemble method was used to improve prediction performance. Four neural networks were trained with the four corresponding balanced datasets.

In the ensemble model, it is important to define a score that combines probabilities from multiple networks. Weighted score ${\tilde P_{ij \cdot }}$ of the ${i}$th image on the ${j}$th skin disease class from an individual network is defined as:



\begin{equation}
{\tilde P_{ij \cdot }} = \frac{{\sum\limits_{\forall k}^{} {w_k^{(j)}{P_{ijk}}} }}{{\sum\limits_{\forall j} {\sum\limits_{\forall k}^{} {w_k^{(j)}{P_{ijk}}} } }}
\end{equation}

where we define a true positive rate of each ${j}$th class of the ${k}$th network as a weight as follows: 

\begin{equation}
{w_{k}^{(j)}}={True\:positive\:rate}
\end{equation}

and $P_{ijk}$ is a probability that the ${i}$th image is classified as the ${j}$th class on the ${k}$th network. The ensemble classifier predicts the disease class with the highest weighted score ${\tilde P_{ij \cdot }}$

\subsection{Implementation} For 100 epochs, the segmentation model was trained using stochastic gradient descent (SGD) optimization with initial learning rate of 0.001 and decaying of 0.9 at every 10 epochs. All the four classification models were trained with the same pre-trained network, and they had the same SGD hyperparameters, but for 50 epochs. 

We developed neural networks with Keras 2.1.4 on Tensorflow 1.4.0 backend, preprocessed images with cv2 in Python 2.7.6. R 3.3.3 was used to generate randomly balanced samples from the training dataset and calculate weighted scores of the ensemble model. Codes for our analysis pipeline are available on \url{https://github.com/YeongChanLee/WonDerM\_ISIC2018\_SkinLesionAnalysis}

\section{Result} The confusion matrices with normalization of classification models over the seven classes are shown in Figure 4. The average balanced accuracy of classification models is 0.836$\pm$0.015 in the development set and 0.840$\pm$0.020 in the validation set. An ensemble of the four classification models showed the improved validation score compared to individual classification models. The ensemble of models achieved an accuracy of 0.899 and 0.785 in the validation set and test set, respectively.

\begin{figure*}
	\centering
	\includegraphics[width=11cm]{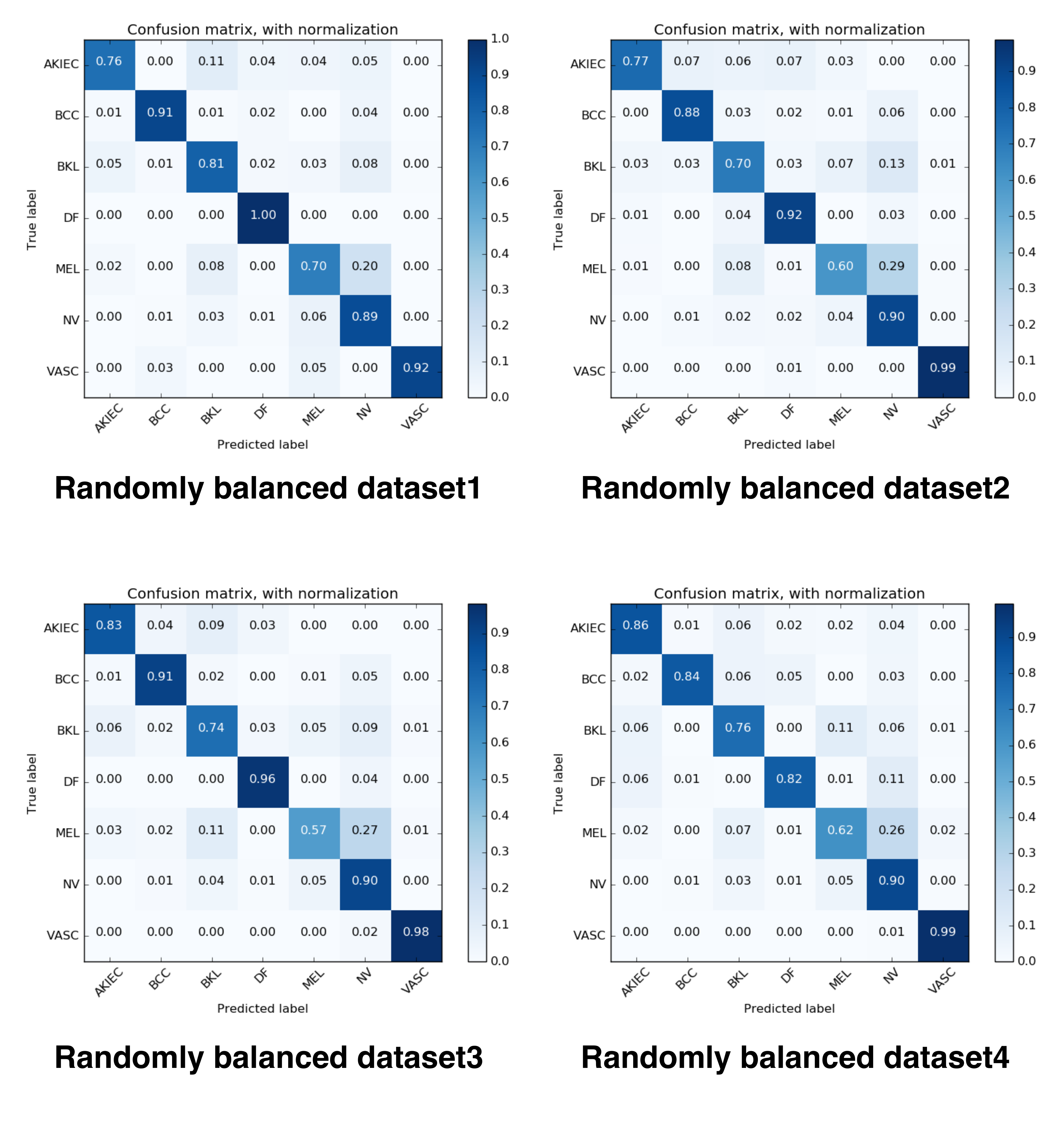}
	\caption{Confusion matrix comparison among classification models.}
\end{figure*}

\section{Discussion} The ISIC Skin Image Analysis Challenge@MICCAI 2018 is a competition for skin lesion analysis of dermoscopic images from the HAM10000 dataset. We suggest the WonDerM pipeline for the task 3 of the challenge that is to classify seven classes of skin disease. The ensemble model from the WonDerM achieved an accuracy of 0.899 and 0.785 in the validation set and test set, respectively.

The major problem by class imbalance is that a model trained using an imbalanced dataset might be biased towards a particular class with larger number of images. Considering that the given training dataset was highly imbalanced, we used a balanced sampling method and ensembled multiple networks. The proposed weighted score of the ensemble was also useful for overcoming the class imbalance problem by weighting prediction probabilities for certain classes that had higher true positive rates in the network. However, the score may be also prone to weighting a certain class with an extremely higher true positive rate than other classes.

The proposed WonDerM pipeline showed good performance to classify multiple skin diseases, which can be applied in other medical imaging-based predictions. We demonstrated that the WonDerM can be adapted if the additional source used to transfer knowledge (ex. semantic segmentation) has more information than the target task (ex. object detection). Further research is required to develop an improved model with clinical information such as age, sex and family history.

\section{Acknowledgement} This study was supported by a grant funded by SKKU Convergence Institute for Intelligence and Informatics (SCI-Cube), Sungkyunkwan University.



%





\ifCLASSOPTIONcaptionsoff
  \newpage
\fi

\bibliographystyle{unsrt}
\bibliography{ISIC}

\end{document}